\documentclass[sigconf]{acmart}

\AtBeginDocument{%
  \providecommand\BibTeX{{%
    \normalfont B\kern-0.5em{\scshape i\kern-0.25em b}\kern-0.8em\TeX}}}

\setcopyright{acmcopyright}
\copyrightyear{2023}
\acmYear{2023}
\acmDOI{XXXXXXX.XXXXXXX}

\acmConference[Conference acronym 'XX]{Make sure to enter the correct
  conference title from your rights confirmation emai}{June 03--05,
  2023}{Woodstock, NY}
\acmPrice{15.00}
\acmISBN{978-1-4503-XXXX-X/18/06}

\usepackage{CJKutf8}
\usepackage{latexsym}
\usepackage{graphicx}
\usepackage{sidecap}
\usepackage{caption}
\usepackage{subcaption}
\usepackage{amsmath}
\usepackage{amsfonts}     
\allowdisplaybreaks
\usepackage{amsthm}

\newtheorem{example}{Example}
\usepackage{thmtools}
\usepackage{thm-restate}

\usepackage{esvect}
\usepackage{wrapfig}

\usepackage{multicol}
\usepackage{pifont}
\usepackage{xspace}

\usepackage{natbib}

\usepackage{appendix}

\newcommand{\rightcomment}[1]{\(\triangleright\) {\small \it #1}}
\newcommand{\eqcomment}[1]{\addtocounter{equation}{1}\tag*{\rightcomment{#1}\quad(\theequation)}}
\usepackage{suffix}
\WithSuffix\newcommand\eqcomment*[1]{\tag*{\rightcomment{#1}}}
\usepackage{setspace}
\usepackage{hyperref}

\usepackage{mdframed}

\usepackage{xcolor}  %

\usepackage{bm}
\renewcommand{\vec}[1]{{\boldsymbol{\mathbf{#1}}}}   %

\usepackage{xspace}

\usepackage{url}            %
\usepackage{booktabs}       %
\usepackage{nicefrac}       %
\usepackage{microtype}      %
\usepackage{xcolor}         %

\usepackage{microtype}

\usepackage{tcolorbox}
\usepackage{soul}
\definecolor{aigold}{RGB}{255, 249, 205} 
\definecolor{aigreen}{RGB}{166, 255, 193}
\definecolor{humanpurple}{RGB}{235, 222, 240}
\definecolor{commentgray}{RGB}{86, 101, 115}

\definecolor{aired}{RGB}{255, 218, 205} 
\newtcolorbox{examplebox}[1][]{%
  colback=white, 
  colframe=blue!75!black, 
  fonttitle=\bfseries,
  title=Example:,
  #1
}

\usepackage{mathtools, cuted}
\usepackage{tabularx, booktabs}
\newcolumntype{C}{>{\centering\arraybackslash}X}
\newcolumntype{R}{>{\raggedleft\arraybackslash}X}
\newcolumntype{S}{>{\raggedleft\arraybackslash\hsize=.5\hsize}X}
\usepackage{url}
\usepackage{xspace}
\usepackage{array}
\usepackage{enumitem}
\usepackage{cases}
\usepackage{mathtools}
\usepackage{empheq}
\usepackage{bbm}
\usepackage{footnote}

\newcommand{\codefont}{\fontfamily{lmtt}\selectfont}
\usepackage{listings}
\usepackage{parcolumns}
\lstdefinestyle{datalogstyle}{
	basicstyle={\codefont\small},  %
	xleftmargin={6pt},
        columns=flexible,
        breakindent=0pt,
        breaklines=true, 
	frame=tb,
	stepnumber=1,
	firstnumber=1,
	numberfirstline=true,
	tabsize=2,
	extendedchars=true,
	breaklines=true,
	columns=fullflexible,
	keepspaces=true,
	escapeinside={@}{@},
	firstnumber=last,
	captionpos=b, 
	commentstyle=\color{black!65},
	numberstyle=\tiny\color{black!65},
	stringstyle=\color{codepurple},
	breakatwhitespace=false, 
	keepspaces=true,              
        mathescape=true, 
	numbersep=5pt,                  
	showspaces=false,                
	showstringspaces=false,
	showtabs=false,
	aboveskip={0.8\baselineskip},
	belowskip={0.2\baselineskip},
}
\usepackage[noabbrev,capitalize]{cleveref} %
\crefname{equation}{equation}{equations}   %
\crefname{section}{section}{sections}      %
\crefname{lstlsting}{listing}{listings}   
\crefname{assumption}{assumption}{assumptions}
\crefname{line}{line}{lines}   %

\let\frac=\tfrac

\renewcommand{\vec}[1]{{\boldsymbol{\mathbf{#1}}}}   %

\newcommand{\set}[1]{\mathcal{#1}}

\usepackage{pifont}  %
\usepackage{makecell}
\newcommand{\cmark}{\textcolor{green}{\ding{51}}}%
\newcommand{\xmark}{\textcolor{red}{\ding{55}}}%

\usepackage{verbatim}
\usepackage{color}

\hypersetup{
    linkbordercolor=blue,
    urlbordercolor=pink,
    citebordercolor=green,
}

\begin{document}

\title{WeaverBird: Empowering Financial Decision-Making with\\ Large Language Model, Knowledge Base, and Search Engine}


\author{Siqiao Xue$^{{\diamondsuit},{*}}$, Fan Zhou$^{{\diamondsuit},{*}}$, Yi Xu$^{\diamondsuit}$, Ming Jin$^{\clubsuit}$, Qingsong Wen$^{\heartsuit}$, Hongyan Hao$^{\diamondsuit}$, Qingyang Dai$^{\diamondsuit}$ and Caigao Jiang$^{\diamondsuit}$, Hongyu Zhao$^{\diamondsuit}$, Shuo Xie$^{\diamondsuit}$, Jianshan He$^{\diamondsuit}$, James Zhang$^{\diamondsuit}$, Hongyuan Mei$^{\spadesuit}$}
\affiliation{${\diamondsuit}$Ant Group, ${\clubsuit}$Monash University, ${\heartsuit}$Alibaba DAMO Academy, $^{\spadesuit}$TTIC \country{ }} 
\email{{siqiao.xsq,hanlian.zf,haolin.xy}@alibaba-inc.com,ming.jin@monash.edu, qingsongedu@gmail.com}
\email{hongyuan@ttic.edu}

\renewcommand{\shortauthors}{Xue et al.}

\begin{abstract}
We present WeaverBird, an intelligent dialogue system tailored for the finance sector. Our system harnesses a large language model of GPT architecture that has been tuned using extensive corpora of financial texts. As a result, our system possesses the capability to understand complex financial queries, such as ``How should I manage my investments during inflation?'', and provide informed responses. Furthermore, our system integrates a local knowledge base and search engine to retrieve relevant information. The final system responses are conditioned on the web search and include proper references to the sources, thus enjoying enhanced credibility. 
Comparative evaluations on a wide range of financial question-answering highlight our system's superior performance over other models.
Experience WeaverBird for yourself through our live demo at {\small\url{https://weaverbird.ttic.edu}} and view our concise 2-minute video at {\small\url{https://www.youtube.com/watch?v=yofgeqnlrMc}}. Our code has also been made publicly available at {\small\url{https://github.com/ant-research/fin_domain_llm}}.
\end{abstract}

\ccsdesc[500]{Social and professional topics~Computing and business}
\ccsdesc[500]{Computing methodologies~Natural language processing}

\keywords{large language models, finance, dialogue system}


\maketitle
\def\thefootnote{\text{*}}\footnotetext{These authors contributed equally to this work.}
\renewcommand{\thefootnote}{\arabic{footnote}}
\section{Introduction}\label{sec:introduction}
Everyone faces financial decisions from time to time. From questions like ``What does the earnings report indicate for stock X?'' to ``How should I navigate my investments during inflation?'', many individuals prefer consulting financial professionals. Yet, not everyone has the means to access such expert services.

In this paper, we introduce WeaverBird\footnote{This name embodies our vision for WeaverBird, symbolizing our commitment to delivering comprehensive and intelligent support in the finance domain. It represents the intricate and skillful weaving of financial knowledge and information, empowering users with a robust and reliable resource.}, an intelligent dialogue system tailored for the financial sector. 
Our primary objective is to bridge the disparity in accessing financial advisory services and to democratize financial expertise. Acting as an intelligent financial assistant, WeaverBird is adept at understanding complex economic and financial queries, offering financial perspectives, and enabling users to make informed financial decisions. Through the launch of our system, we aspire to ensure that people across varied financial backgrounds gain access to trustworthy financial advice.

\begin{table*}
\scalebox{0.9}{
\begin{tabular}{ l c c  c  c c   }
\toprule
                      & \makecell{WebGLM~\cite{liu2023webglm}}
                      & \makecell{BloombergGPT~\cite{wu2023bloomberggpt}}
                      & \makecell{FinGPT~\cite{yang2023fingpt}}
                      & \makecell{FinChat~\cite{finchat}} 
                        & \makecell{WeaverBird} \\
\midrule
Finance-domain specific LLM & \xmark & \cmark & \cmark & \cmark & \cmark \\
Finetuned on bilingual financial data     & \xmark       & \xmark       & \cmark  & \xmark & \cmark                  \\
Local knowledge base querying  & \xmark  & \xmark    & \cmark     & \cmark    & \cmark     \\
Web search querying  & \cmark  & \xmark    & \xmark     & \cmark    & \cmark     \\
Efficiency-optimized search on multi-source knowledge  & \xmark  &  \xmark & \xmark & N/A    & \cmark      \\
Temporal reasoning optimized  & \xmark & \xmark & \xmark   & N/A  & \cmark      \\
Open access to training data $\&$ code   & \xmark & \xmark       & \cmark  & \xmark      & \cmark                   \\
Open access to deployed chatbot   & \xmark  & \xmark      & \xmark       & \xmark         & \cmark          \\
\bottomrule
\end{tabular}
}
\caption{Comparative summary of competing approaches on various dimensions.}
\label{fig:comparison}
\vspace{-2mm}
\end{table*}

Another key motivation for developing a finance dialogue system is to support future research in this area. 
A comparison of our WeaverBird system to other competitive approaches is summarized in \cref{fig:comparison}.
Throughout the development process of our system, we have amassed extensive collections of finance-centric text documents, which have been leveraged in fine-tuning language foundation models. This resource will be available to the research community for future exploration and analysis. Notably, our collection features a substantial corpus of Chinese financial documents, addressing the gap in the predominantly English-based finance corpora and offering a valuable addition. 
This equips our WeaverBird system with a distinctive edge in processing Chinese queries and delving into discussions concerning Chinese firms and equities.
Furthermore, we have constructed a collection of challenging financial question-answer pairs, which can serve as a benchmark for evaluating the performance of future models and systems. Each question stems from a selected document, such as an industry research report. Consequently, answering these questions correctly requires the discovery of the specific document or equivalent information from the local knowledge base or across the internet. Our WeaverBird system harnesses the latest research advancements in large language models, enhanced by our novel training and inference methods, which we will elaborate on in \cref{sec:overview}. This system sets a robust baseline for evaluating future models and systems in the finance domain.

An additional compelling motivation is the opportunity to leverage the knowledge and assets acquired during the development of our system to create intelligent assistants in broader domains, such as education and healthcare.
\section{System Design}\label{sec:overview}
\begin{figure*}[t]
    \centering
    \includegraphics[width=0.9\linewidth]{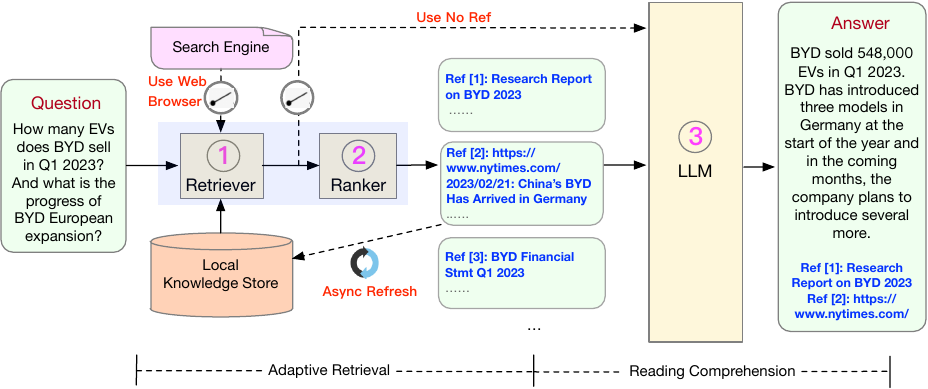}
    \caption{An illustration of WeaverBird that answers a financial query by intelligent search and generation.}
    \label{fig:overview}
\end{figure*}

When presented with a natural language prompt, such as ``How many EVs did BYD sell in Q1 2023?'', our system employs a two-phase process. First, it transforms the prompt into queries suitable for the knowledge base and search engine. Next, it analyzes the retrieved information, identifies the most relevant pieces, and generates a comprehensive natural language response, with proper references to the original sources. The Internet's vast resources enable us to craft responses and insights based on up-to-date information, enhancing the adaptability and reliability of our approach. The overall pipeline of WeaverBird is depicted in \cref{fig:overview}.  While it builds upon the general framework of LangChain~\cite{langchain,langchainglm}, our system integrates our novel training and inference techniques, which significantly enhance its overall performance and efficiency. In this section, we delineate the design of each phase, including the model architecture as well as training and inference paradigms.

\subsection{Efficiency-Optimized Search and Update}\label{sec:search}\label{sec:retrieval}
The first phase is to retrieve information from our local knowledge base and the broader internet (i.e., the first step in \cref{fig:overview}). Our knowledge base $\set{K}$ is a collection of financial documents $\vec{d}^{\text{loc}}_1, \ldots, \vec{d}^{\text{loc}}_N$ where the number of documents $N$ is large. Following \citet{langchain}, we split each document $\vec{d}_n$ into multiple paragraphs $\vec{p}^{\text{loc}}_{n,1},\ldots, \vec{p}^{\text{loc}}_{n,M_n}$, where $M_n$ (and $m$ below) denotes the index of paragraph for the $n$-th document, and embed each paragraph into a multidimensional embedding $\vec{e}^{\text{loc}}_{n,m}$ through a neural encoder $f_{\text{key}}$. When a language query $\vec{x}$ comes, it is embedded into a vector $\vec{q}$ through another neural encoder $f_{\text{query}}$ and we retrieve the top $K$ relevant paragraphs from the knowledge base according to their cosine similarities, i.e., $\vec{q}^{\top}\vec{e}/\|\vec{q}\|\|\vec{e}\|$, where $K$ is a hyperparameter.

If the cosine similarity of the most relevant retrieval is larger than a predefined threshold $c\in(0,1)$, we skip the internet search and proceed directly to the response generation phase that we present in \cref{sec:gen}. 
Otherwise, we start an internet search, which is slower than a local search but may yield higher-quality results. 
This efficiency-optimized design is not in the original LangChain pipeline and we propose it to strike a balance between performance and efficiency.\footnote{Internet search is also not in the original LangChain, but it is in the latest release, concurrent with our work.}
When the local knowledge base already provides highly relevant information, we can avoid the need to search the internet. 
Otherwise, we invest additional effort to retrieve internet-based information for enhanced quality.%

When an internet search\footnote{The search engine we use is Google Search.} is used, we select the highest-ranked $N$ documents $\vec{d}^{\text{web}}_1, \ldots, \vec{d}^{\text{web}}_N$, split each document into paragraphs $\vec{p}^{\text{web}}_{n,1},\ldots, \vec{p}^{\text{web}}_{n,M_n}$, and embed each paragraph into a vector $\vec{e}^{\text{web}}_{n,m}$ through the neural encoder $f_{\text{key}}$ that we have used for the knowledge base. 
Then we select the top $K$ relevant paragraphs according to their cosine similarities. 
Combined with the local retrievals, we end up with $2K$ paragraphs and proceed to the next phase. 

If any of the web paragraph exhibits a cosine similarity greater than the threshold $c$, our system will automatically add the entire document to which it belongs into our local knowledge base. 
Users can also manually add any online document. 
This update ensures that future queries, whether identical or related, will receive more efficient responses. 
As updates occur, the knowledge base will expand, but the local search is still significantly more efficient than a web search. 

\subsection{Learning to Embed and Search}\label{sec:learn}
We confidently consider a higher similarity to signify a more relevant paragraph due to the
training of the encoders $f_{\text{key}}$ and $f_{\text{query}}$. Their optimization involves maximizing a well-defined objective:
\begin{align}
    \ell 
    = 
    \vec{q}^{\top} \vec{e}_{0}
    - \log \sum_{i=0}^{I} \exp\left( \vec{q}^{\top} \vec{e}_{i} \right),
\end{align}
where $\vec{e}_{0}$ is the embedding of the paragraph known to contain relevant information for the query, and the other $I$ embeddings $\vec{e}_{1}, \ldots, \vec{e}_{I}$ belong to a set of negative paragraphs (see \cref{sec:qgen} for how they are selected). 
Intuitively, we want the dot products $\vec{q}^{\top} \vec{e}$ to be relatively large for the query-paragraph pairs that are actually relevant. Our encoders use Multilingual-E5-base model architecture~\citep{wang2022text}. 

Once we have trained the encoders, we could determine the threshold $c$ by computing the cosine similarities for a holdout set of query-paragraph pairs and analyzing the histogram.\footnote{We experimented with dot product and cosine similarity and found that the latter yields a better overall performance even though our training objective has used the former}. In practice, we choose the 1\% quantile to be our threshold $c$. 

Another question is where our query-paragraph pairs come from. We have collected a corpus of query-paragraph pairs by leveraging financial documents and question generation methods, which we will outline in \cref{sec:data}.

\subsection{Generation by Large Language Models}\label{sec:gen}
In this phase, our system analyzes the search results, fills an appropriate prompt template and generates a response. We finally determine the strategies as follows:


\paragraph{Ranker.} The system firstly first ranks all the $2 K$ search results (or $K$ results if the internet search has been skipped as discussed in \cref{sec:retrieval}) based on their cosine similarities with the query (i.e., the second step in \cref{fig:overview}).

\paragraph{Prompt Formulation with Temporal Context.} After receiving the ranking results from the previous step, we plug the top $J$ (where $J < K$) results into the context part of the predefined prompt template. Since the knowledge of finance domain is dynamically evolving, shown in \cref{lst:en_template}, we ground each retrieved result and question with temporal facts (e.g., published date and question date) to facilitate the language model to adapt to time-evolved knowledge, possibly making the response more meaningful and accurate~\cite{jin2023large}. \\

Having temporal metadata for questions and retrieved results allows us to ask questions with relative time specifications (e.g., ``3
months ago''), which are often used in finance queries. Moreover, it enables us to ingest new
knowledge periodically, in a \emph{streaming} setup, and evaluate on questions asked during that period~\citep{streamingqa2022}. \\


\begin{lstlisting}[caption={Prompt templates for LLM.},label={lst:en_template}]
@
\sethlcolor{aigold}  
\hl{
English template
}
@

Context information:
{CONTEXT_DATE_1:CONTEXT_RETRO_1} 
@
\vdots
@
{CONTEXT_DATE_2K:CONTEXT_RETRO_2K} 

The current date is {QUESTION_DATE}. Based on the given information, please provide a concise and professional response to the user's question. If there are multiple questions in a query, please answer all of them. If the user's question includes keywords like 'recent' or 'latest' to indicate a recent time frame, pay attention to the correspondence between the current date and the date of the information. If a clear answer cannot be determined, respond with "Unable to answer the question based on the information provided". You MUST respond in the same language as the question!

The question is: {QUESTION}.
\end{lstlisting}
\vspace{-0.3cm}
\paragraph{Response Generation.}  The large language model organizes the information to form a fluent and informed response, corresponding to the third step in \cref{fig:overview}. The response also includes accurate references to the sources of the information. As an option, our system can further provide a list of citations for the remaining $(K - J)$ paragraphs at the end of the response, to complement the main response and provide users with additional recommended reading materials. \\

Our large language model has undergone adaptation to the finance domain, thus being able to skillfully process the input information and generate appropriate responses. We started with a pre-trained GLM~\citep{du2022glm,zeng2022glm} that has been pre-trained using extensive English and Chinese corpora. Then we fine-tuned this model using our collected Chinese and English financial documents as well as expert-annotated financial query-response pairs, which we introduce in \cref{sec:data}. 
\section{Data Collection and Utilization}\label{sec:data}
In this section, we introduce our corpora of financial documents and their collection process, as well as how we used them to build the local knowledge base and generate training and evaluation query-paragraph pairs. 

\subsection{Collection of Financial Documents}\label{sec:doc}
We collected a large set of financial documents from well-known and accredited sources, most of which are global financial institutes. 
Our collection includes 240 million Chinese documents and 25 million English documents. 
They were all published between 2010-Jan-01 and 2023-June-30, and cover diverse topics such as macroeconomics and fixed income. 
\cref{tab:distribution_publisher_fin_docs} shows a break-down into the information sources, and \cref{tab:distribution_report_cate_fin_docs} shows a break-down into the topics. 

\begin{table}
\centering
\begin{small}
\begin{tabular}{lrlr}
\toprule
Chinese docs &    & English docs & \\
\midrule
 CICC &  $3.5\%$  & Deutsche Bank  & $18.5\%$ \\
 CITIC Sec. &  $3.1\%$  & CICC  & $17.3\%$ \\
 Industrial Sec. &  $2.9\%$  & HSBC  & $11.8\%$ \\
 Haitong Sec. &  $2.8\%$  & CITIC Sec.  & $6.4\%$ \\
 Guotaijunan Sec. &  $2.7\%$  & Guotaijunan Sec. & $5.6\%$ \\
 Huatai Sec. &  $2.1\%$  & Kim Eng Sec. & $4.8\%$ \\
 Guangfa Sec. &  $2.0\%$  & BNP Paribas & $3.9\%$ \\
  Changjiang Sec. &  $2.0\%$  &  Industrial Sec. & $3.7\%$ \\
  CMS &  $2.0\%$  &  Shenwan Hongyuan & $3.7\%$ \\
  Founderfu &  $1.9\%$  &  UBS & $3.1\%$ \\
\bottomrule
\end{tabular}
\caption{Top 10 sources of our financial documents.}
\label{tab:distribution_publisher_fin_docs}
\end{small}
\end{table}

\begin{table}
\centering
\begin{small}
\begin{tabular}{lrlr}
\toprule
Chinese docs &    & English docs  & \\
\midrule
 Financial Engr. &  $25.8\%$  & HK Stocks  & $34.9\%$ \\
 Company &  $20.5\%$  & Company  & $21.7\%$ \\
 Industry &  $20.5\%$  & Industry  & $11.8\%$ \\
 Strategies &  $6.9\%$  & Macro econ  & $18.5\%$ \\
 HK Stocks &  $5.5\%$  & Strategies & $5.6\%$ \\
 Morning report &  $4.9\%$  & Morning report & $4.8\%$ \\
 Macro econ &  $4.8\%$  & Bond & $3.9\%$ \\
  China GEM &  $4.5\%$  &  China GEM & $3.7\%$ \\
  Bond &  $4.1\%$  &  Financial Engr. & $0.1\%$ \\
  Fund &  $1.5\%$  &  Fund & $0.1\%$ \\
\bottomrule
\end{tabular}
\caption{Top 10 topics of our financial documents.}
\label{tab:distribution_report_cate_fin_docs}
\end{small}
\end{table}

For each document, we save into a text file the concatenation of its release time, title, summary, and other metadata (such as topic). By manually analyzing a sampled set of documents and consulting domain experts, we recognized that every document has a summary section that includes nearly all the important information of the document. 
Therefore, we did not save the full content, which will consume a significantly larger amount of disk space and memory but contribute minimal additional information. 
After processing, the average token count for Chinese documents is 5000 and that for English documents is 2000. 
Below are two examples of processed documents:  
\begin{example}[Chinese]
\begin{CJK}{UTF8}{gbsn}
2023-04-03 00:00:00;海底捞(6862.HK)：\\2H22净利率7.5\%; 
门店重启稳步推进;海底捞发布2022年度业绩\ldots (full content omitted);
港股;个股;公司定期报告评述;股票;中国香港.
\end{CJK}
\end{example}
\begin{example}[English]
2023-02-07 00:00:00; 
Feb 223 ECB monetary policy 
meeting commentary: ECB maintains pace of rate 
hikes without reinforcing\ldots (full content omitted);
macro;oversea;finance;economics research;economics;europe. 
\end{example}

The primary use of our large-scale data collection is to build a local knowledge base. 
As mentioned in \cref{sec:retrieval}, we split each document $\vec{d}^{\text{loc}}_n$ into smaller chunks $\vec{p}^{\text{loc}}_{n,1},\ldots, \vec{p}^{\text{loc}}_{n,M_n}$ (about 250 tokens per chunk), and embed each chunk into a vector $\vec{e}^{\text{loc}}_{n,m}$ with our trained neural encoder $f_{\text{key}}$. 
We implemented this vector-based knowledge base and its retrieval method using the Faiss~\citep{johnson2019billion}, a library highly optimized for efficient similarity search and clustering of dense vectors. 

\subsection{Collection of Query-Response Pairs}\label{sec:instruction}

Another important use of our data collection is to create query-response pairs for tuning the language model. This step is crucial for the performance of our system: recent research has shown that language models tuned with query-response pairs have demonstrated superior performance in answering open-domain questions and engaging in conversations~\citep{peft}. 
We collected four kinds of query-response pairs, three of which were created from our collection of financial documents. 
\cref{tab:distribution_pairs} shows a breakdown of the pairs into these four categories. 
In each category, 60\% of the pairs are in Chinese while the rest are in English. 

\paragraph{Generation}
For each document, we can construct a query-response pair that tunes the language model to perform content generation. 
To create the query, we fill the template ``Please write a [TYPE] on [TOPIC] with the title [TITLE]'' where TYPE, TOPIC, and TITLE can be found in the corresponding text file. 
The SUMMARY part of the file is used as the response. 
An example of query is ``Please write a review report on China economics with the title CHINA POLICY: POLITBURO MAINTAINS PRO-GROWTH STANCE'' and its response is ``The politburo indicates continuous pro-growth policy without big stimulus\ldots'' 

\paragraph{Summarization}
The reverse of a content generation pair can be used as a content summarization pair. 
Now the query will be created by filling the template ``Please summarize the following [TYPE]: [SUMMARY]'' and the response is the TITLE. 
An example of query is ``Please summarize the following report: NIO's 4Q22 earnings missed our prior forcast\ldots'' and its response is ``NIO INC.\@ (NIO.US): 4Q22 miss''.

\paragraph{Close-book QA}
Given a document, we would like the model to answer any questions that could be answered by directly identifying the relevant information from the document. 
This is the focus of the close-book QA research~\citep{roberts-etal-2020-much}. 
To achieve this goal, we need a specific kind of query-response pairs: query is the question and the document; response is the answer. 
We created such pairs using the question generation method proposed by~\citet{rajpurkar2016squad}. 
Precisely, given a document, we used a pre-trained question generation model to identify some interesting information in the document and then generate a question whose answer should be that piece of information. 

\paragraph{Open QA}
We would also like the system to answer open questions: unlike close-book QA, the evidence is not available in the context. 
We created this kind of Chinese question-answer (i.e., query-response) pairs by scrapping well-known online QA forums such as Zhihu and SinaFinance. 
On such forums, users post questions, tag questions, answer questions, and upvote or downvote answers. 
We identified 6,000 highly recognized questions with the ``finance'' tag; with each question being the query, we took the answer with the most upvotes to be its response. 
We then randomly sampled 4,000 Enligsh open QA questions from the English finance QA dataset FinQA~\citep{wealth-alpaca}. %

\begin{table}[t]
\centering
\begin{small}
\begin{tabular}{>{\centering\arraybackslash}p{0.22\linewidth} >{\centering\arraybackslash}p{0.05\linewidth} >{\centering\arraybackslash}p{0.1\linewidth} p{0.5\linewidth}}
\toprule
Use case &  \%  & \# Pairs   & Source    \\
\midrule
Generation & 38\% &  13000   & Collected financial documents \\
Summarization  & 20\% & 7000  &  Collected financial documents \\
Close-book QA  &  12\% & 4100 & Collected financial documents \\
Open QA  &  30\% & 10000 &  Chinese: collected from online QA forums (e.g., Zhihu and SinaFinance) \\
 &     & & English: sampled from FinQA \citep{wealth-alpaca}   \\
\bottomrule
\end{tabular}
\caption{Distribution of use case categories of the query-response pairs.}
\label{tab:distribution_pairs}
\end{small}
\vspace{-5mm}
\end{table}

\subsection{Data Collection for Training Encoders}\label{sec:qgen}
Recall from \cref{sec:learn} that we need query-paragraph pairs to train the key and query encoders $f_{\text{key}}$ and $f_{\text{query}}$. 
The close-book QA subset available in \cref{sec:instruction} provides a direct resource to fulfill this requirement. 
First, we categorized all the close-book query-response pairs according to their topics. 
Then we manually reviewed the data of each topic and selected a total of 1,000 pairs while ensuring a comprehensive coverage of all topics. 
The selection of each pair was based on our anticipated likelihood that someone would ask this question during interaction with the demo. 
We used these 1,000 query-response pairs as the positive pairs in training the encoders. 
For each of them, we randomly sampled five negative responses from the entire pool of paragraphs. 
\section{Related Work}\label{sec:related}
Large language models such as ChatGPT~\cite{brown-2020-gpt} and GPT-4~\cite{gpt4} have showcased their remarkable capabilities in engaging in human-like communication and understanding complex queries, bringing a trend of incorporating LLMs in various fields~\cite{anil2023palm,gunasekar2023textbooks,jin2023timellm}.
Recent advancements have further enhanced these models by incorporating external tools, enabling them to search for relevant online information~\citep{nakano2021webgpt}, utilize tools~\citep{schick2023toolformer}, and create more sophisticated applications~\citep{langchain}. 
In this paper, we harness these advancements to build an intelligent dialogue system that specializes in the finance domain. 
The most relevant work is BloombergGPT~\citep{wu2023bloomberggpt}, a generative Transformer trained with large corpora of finance-related text. 
While BloombergGPT has only been trained with English corpora, our system is also carefully tailored to handle Chinese, a language spoken by one-fifth of the global population. 
In addition, BloombergGPT is only trained to predict the next token given the context but not to engage in conversations like our WeaverBird system; nor is it integrated with knowledge bases or search engines. 

Our work aligns with research on retrieval-based language generation, which involves enhancing language models by incorporating external datastores. 
This area has seen advancements in new model architectures that condition the next token generation on retrieved text~\citep{khandelwal2019generalization,wu2022memorizing,borgeaud2022improving}, as well as developments in training and inference methods aimed at improving the quality of retrieval~\citep{zhong2022training} and the performance in downstream tasks such as question answering~\citep{karpukhin2020dense,izacard2022few,shi2022nearest,menick2022teaching} and code generation~\citep{zhou2022docprompting}. 
Our work aims to deliver a system that has superior performance in answering finance-related questions. 
It uses the standard GPT architecture, and augments the inputs (i.e., queries) with retrieved information (i.e., online documents or KB entries): this design is similar to the approaches of~\citet{shi2023replug,ram2023context}, but our efficiency-optimized search and update (\cref{sec:search}) is novel.

\begin{figure*}[t]
    \begin{minipage}[t]{0.64\linewidth}
        \includegraphics[width=0.48\linewidth]{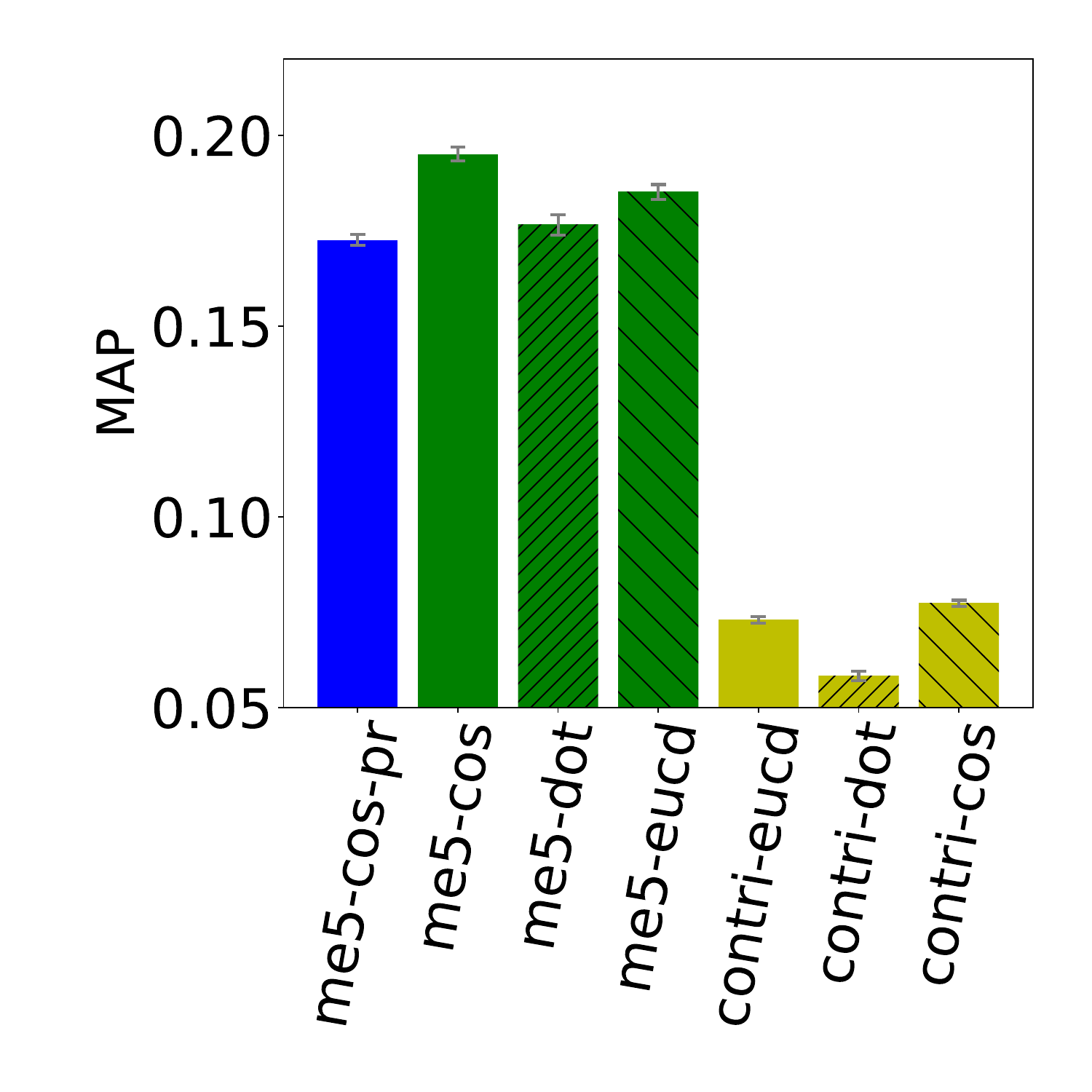}
        ~
        \includegraphics[width=0.48\linewidth]{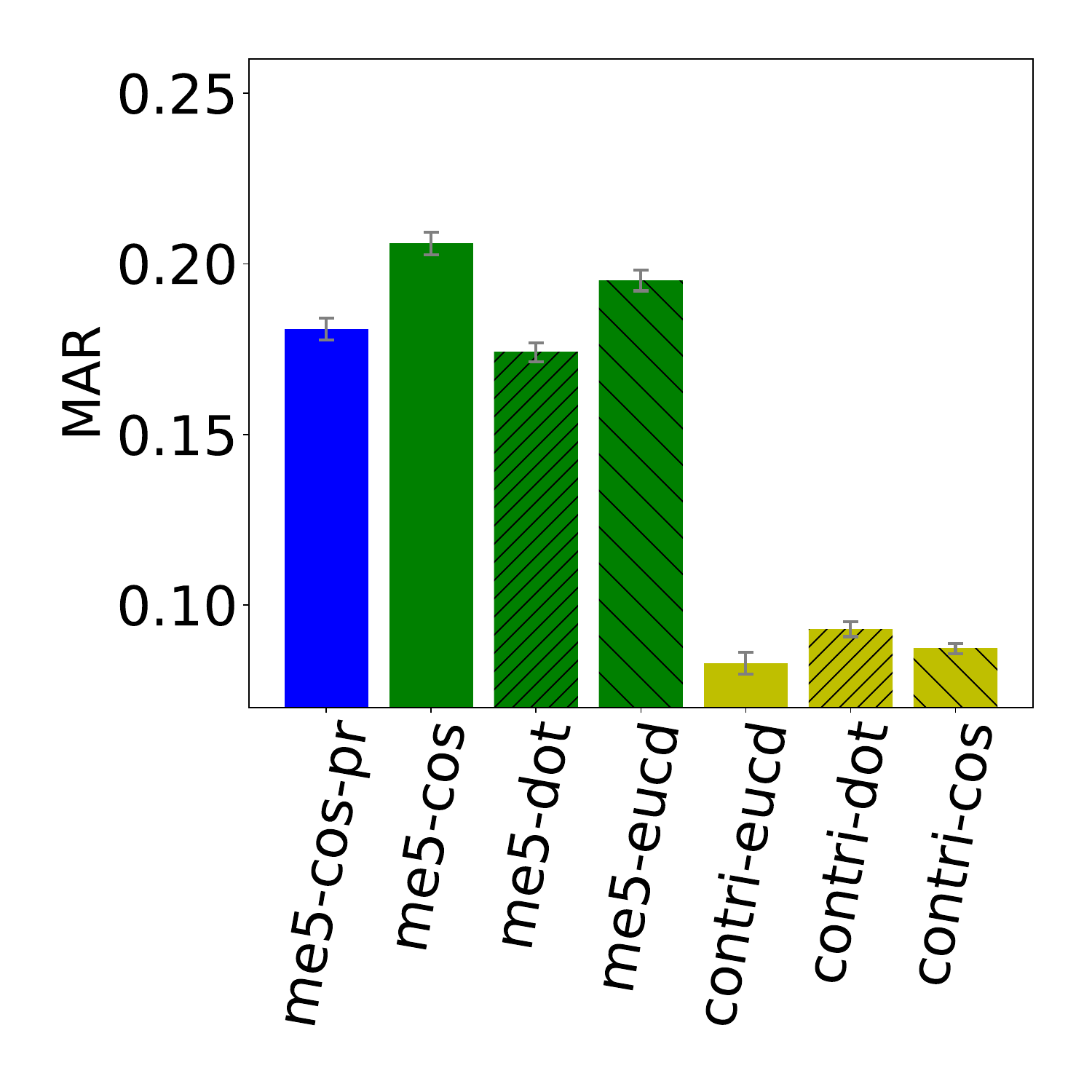}
    \caption{Retrieval performance of all combinations of encoders and similarity scores. From left to right, they are: pretrained ME5 with cosine similarity, trained ME5 with cosine similarity, trained M3E with dot product, trained ME5 with Euclidean distance, trained Contriver with Euclidean distance, trained Contriver with dot product, and trained Contriver with cosine. Pre-trained Contriver is not presented since its performance is very poor.}
    \label{fig:retro}
    \end{minipage}
    \hfill 
    \begin{minipage}[t]{0.31\linewidth}
        \includegraphics[width=\linewidth]{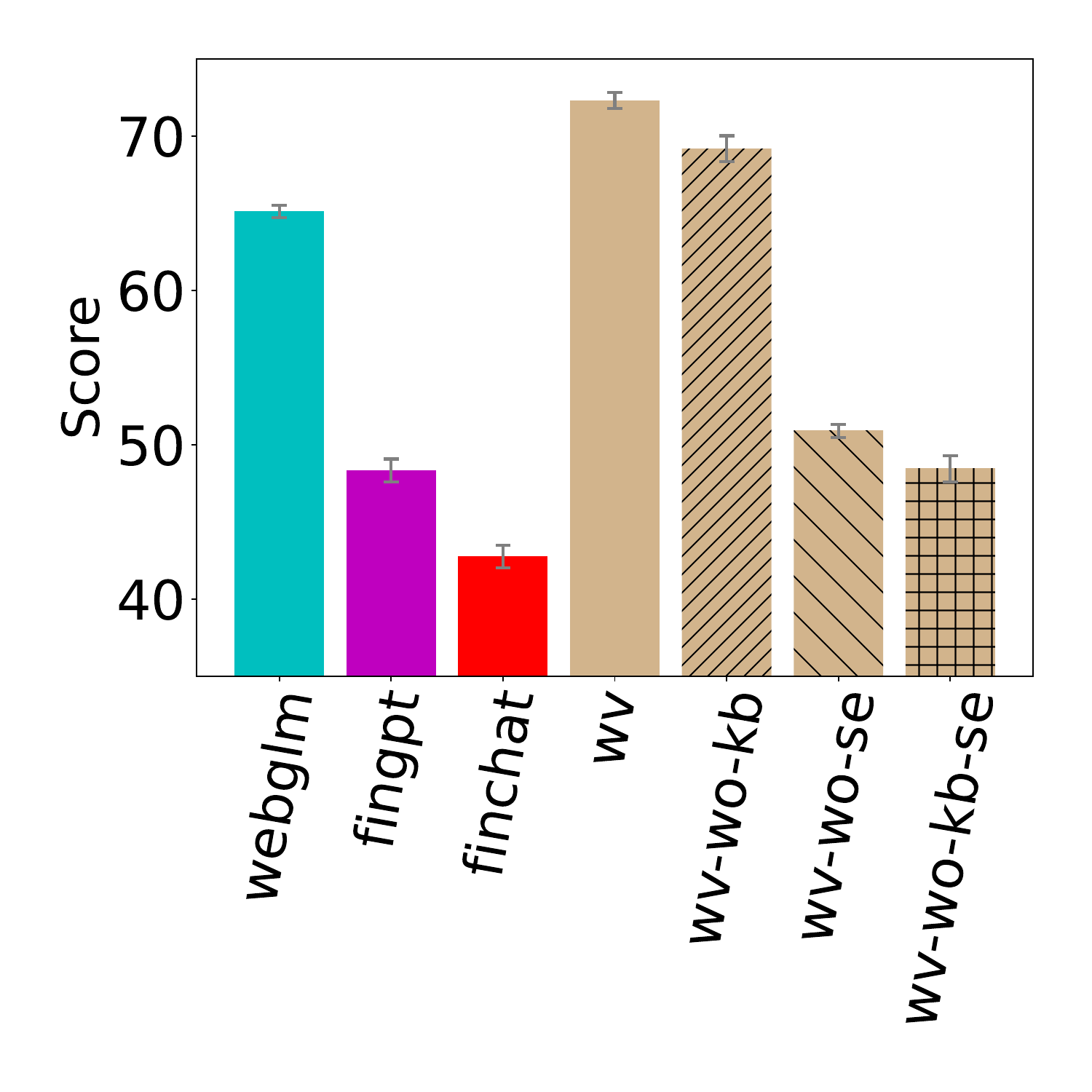}
        \caption{Response quality performance of all methods. From left to right, they are: WebGLM, FinGPT, FinChat, WeaverBird, WeaverBird without knowledge base, WeaverBird without search engine and WeaverBird with neither knowledge base nor search engine.}
    \label{fig:qa_score}
    \end{minipage}
\end{figure*}

\section{Models and Training}
\label{sec:model}

\subsection{Language Model}
\label{sec:llm_model}
\paragraph{Architecture.} The system is based on the pre-trained ChatGLM2-6B~\citep{du2022glm,zeng2022glm}, and we fine-tune it with the data and training strategy specified below. 

\paragraph{Dataset.} We use the collected query-response pairs clarified in \cref{sec:instruction} for training and evaluation. The whole set of pairs are divided into sets of 17,000 training pairs, 3,000 development pairs, and 4,100 test pairs. 

\paragraph{Training.} We apply the LoRA~\citep{hu2021lora} strategy to fine-tune the base model to adapt to the finance domain. The LoRA layer details are as follows: the LoRA rank is 16, the LoRA alpha is 32, the dropout is 0.1, and other settings are default.

\subsection{Encoders}
\label{sec:encoders}
\paragraph{Architecture.} The key and query encoders $f_{\text{key}}$ and $f_{\text{query}}$ are initialized as Multilingual-E5-base (ME5) model architecture~\citep{wang2022text}.

\paragraph{Dataset.}  Recall from \cref{sec:learn} that we need query-paragraph pairs to train the key and query encoders $f_{\text{key}}$ and $f_{\text{query}}$. We use the collected query-response pairs clarified in \cref{sec:gen} for training and evaluation. The chosen pairs were then divided into sets of 700 training pairs, 100 development pairs, and 200 test pairs. 

\paragraph{Training.} We pass query-response pairs to the
model to yield a scalar score for each of the pair
and maximize the scores for the positive pairs while minimizing the scores for the negative pairs with the cross entropy loss.

\subsection{Implementation and Deployment Details.}
\label{sec:implementation}

\paragraph{Knowledge Base and WebUI.} For the implementation of LangChain, we use the code from  the  public  GitHub  repository at {\small \url{https://github.com/imClumsyPanda/langchain-ChatGLM}}~\citep{langchainglm} without any License. For the implementation of WebUI, we use the code from  the  public  GitHub  repository at {\small \url{https://github.com/THUDM/WebGLM}}~\citep{liu2023webglm} with Apache 2.0 License.

\paragraph{Computation Cost.} All the experiments were conducted on a server with 128G RAM, a 64 logical cores CPU (Intel(R) Xeon(R) Platinum 8163 CPU @ 2.50GHz) and 8 NVIDIA A100 80G Tensor Core GPUs for acceleration. During the fine-tuning process on the finance dataset,  we set the   train batch size on each device equal to $8$, the gradient accumulation step equal to 100, the max step equal to $10,000$ and the learning rate equal to 0.0001, and the running time of each step is close 72 seconds, the total running time is close to $200$ hours.

\paragraph{Deployment Details.}
Our system is deployed on a server on Alibaba Cloud with 30G RAM, a 8 logical cores (Intel Xeon(Ice Lake) Platinum 8369B), and a NVIDIA A100 80G Tensor Core GPU. The inference speed depends search speed on the website, searches in local knowledge only take 1 second, and the search on the website takes 10 seconds on average.

\section{Experiments}
\label{sec:experiment}
We present the experiments designed to evaluate the performance of the WeaverBird system, including the effectiveness of the trained encoders (\cref{sec:retrievaleval}) and the generation quality (\cref{sec:geneval}).

\subsection{Paragraph Retrieval Evaluation}
\label{sec:retrievaleval}

Now we show the results of evaluating the trained encoders against pretrained ME5 and trained Contriver-based encoders~\citep{contriever-2022}. 

\paragraph{Experimental Setup.}  We use the test set specified in \cref{sec:encoders} for evaluation. When evaluating a method (i.e., a specific model paired with a specific similarity metric), we take the top $K=5$ highest-ranked retrievals for each given query, and evaluate the mean average precision (MAP) and mean average recall (MAR) on the entire test set of query-paragraph pairs. Intuitively, MAP measures the fraction of the predictions that are correct, while MAR measures the fraction of the actual paragraphs that are covered by the predictions. Higher is better. 


\paragraph{Results.} The results are in \cref{fig:retro}. 
As we can see, trained encoders work significantly better than pretrained encoders, indicating the effectiveness of our training method. 
Moreover, our ME5-based encoders significantly outperform the Contriver-based encoders: it is perhaps because a large portion of our test cases are in Chinese and ME5 is a bilingual model. 
Interestingly, cosine similarity turns out to work the best among all the similarity metrics, even though our training objective uses dot product.

\subsection{Response Quality Evaluation}
\label{sec:geneval}

\paragraph{Experimental Setup}.  We use the test set specified in \cref{sec:llm_model} for evaluation. Since automatic evaluation for complicated natural language QA still remains an open research question, we rate the response by both human expert and GPT-4~\citep{gpt4} due to its efficiency and consistency~\citep{hackl2023gpt4}: each response is given a score in $[0, 100]$ by a financial expert and GPT-4, respectively and the final score is an average of the two. Higher is better. We compare WeaverBird with WebGLM, FinGPT and FinChat.
BloombergGPT~\cite{wu2023bloomberggpt} is non open-sourced so it is excluded from the baselines. Due to the budget constraint,  we evaluate FinChat, a commercial chatbot in finance, on a subset of 300 questions sampled from the test set.


\paragraph{Results.} The strongest results of our WeaverBird system is summarized in \cref{fig:qa_score}. Although WebGLM is also bilingual and empowered with search engine, WeaverBird has superior performance mainly due to its domain-finetuned base model, trained encoders on financial documents as well as optimized retrieval on knowledge base and searched results. FinGPT has lower score than WeaverBird and WebGLM because it has not used multi-source knowledge. FinChat performs poorly because it is not billingual and its knowledge base covers a restricted list of stocks.

\subsection{More Analysis}
\paragraph{Analysis I: Effect of knowledge base and search engine on the response quality.} Firstly, we remove the knowledge base (`wv-wo-kb' in \cref{fig:qa_score}) and this strategy causes a moderate drop in performance. Secondly, we remove the search engine (`wv-wo-se' in \cref{fig:qa_score}) and the performance declines with a notable drop, indicating the web search ability, which introduces the `global knowledge base' on the internet, greatly improves the accuracy of responses. To conclude, the searched results contribute most to the response quality while the local knowledge base contributes moderately.  

\begin{wrapfigure}{L}{0.5\linewidth}
\vspace{-3mm}
\includegraphics[width=4.3cm]{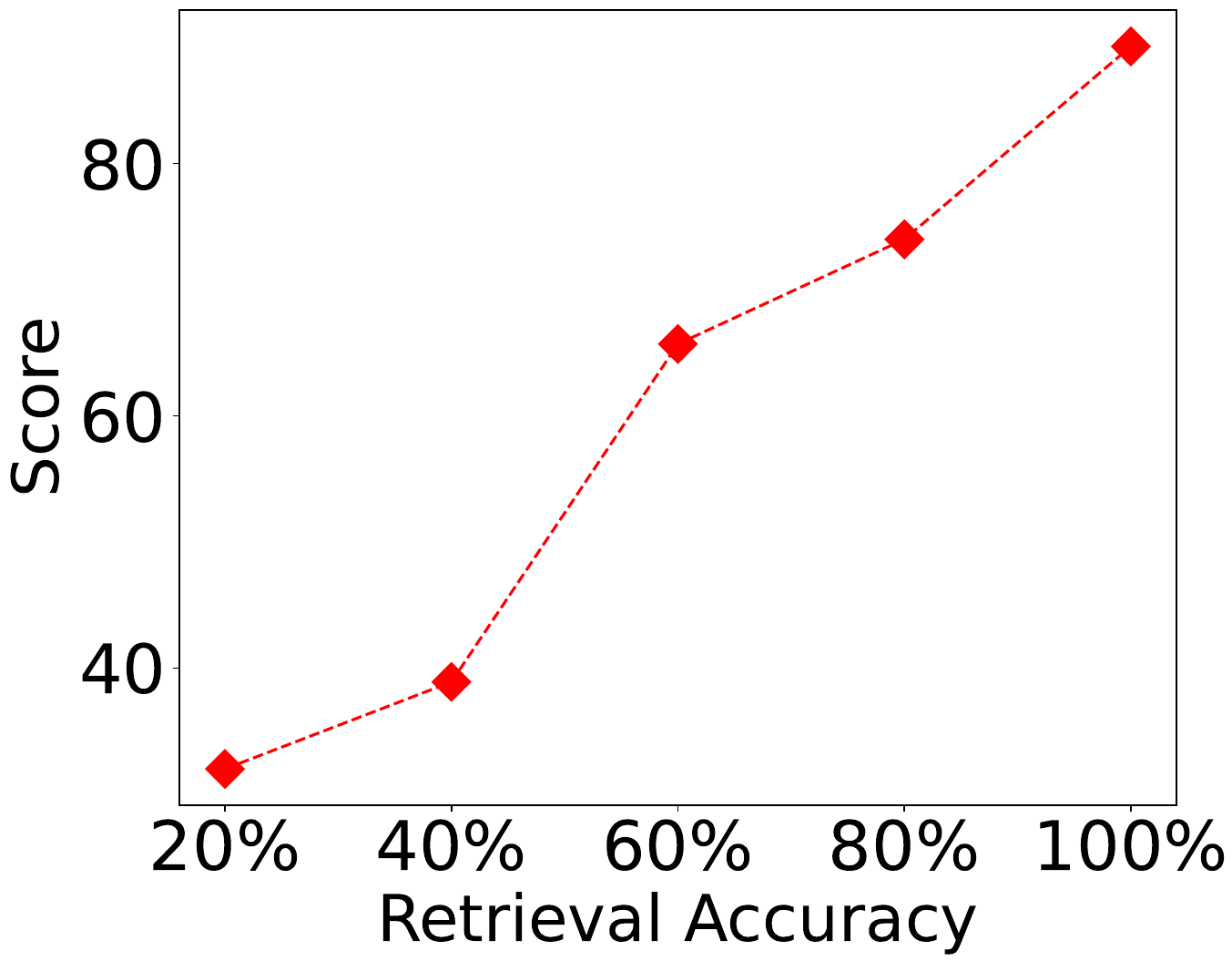}
\caption{The effect of document retrieval accuracy on the response quality of the WeaverBird system.}
\label{fig:ablation_retro}
\vspace{-2mm}
\end{wrapfigure}

\paragraph{Analysis II: Effect of the retriever on the response quality?} We randomly pick up $500$ examples of query-response pairs generated by WeaverBird and group the responses by the accuracies of retrieved documents (annotated by the experts). Not surprisingly, \cref{fig:ablation_retro} shows that higher accuracy of retrievals brings significant improvement on response qualities.

\paragraph{Analysis III: How to correct flawed cases.}
We manually analyze $50$ examples of finance-related query-response pairs annotated by the experts. Of these $50$, $70\%$ cases arrived at reasonable responses. However, we also found some flawed output, most of which can be corrected with one of the following classes of modification:
\begin{itemize}[leftmargin=*]
    \item Increase $K$ to have more searched results; increase $J$ to have more input for LLM; or increase both $K$ and $J$.
    \item The searched results usually consist of a number of relevant paragraphs instead of the full content of the web page. If needed, by using 
    the functionality provided by our system, the user can choose to scrape the site and save the content to the local knowledge base. Given the full text, more relevant paragraphs are retrieved and better responses are possibly generated.
\end{itemize} 
\paragraph{Analysis IV: Qualitative examples of challenging queries.}

In this section, we present two illustrative examples. 
The first example poses a challenging question regarding the holdings of a renowned fund manager. 
Obviously, WeaverBird outperforms the competitors by delivering a reasonably accurate response, accompanied by citations to credible sources. \\

\begin{lstlisting}[caption={About company dividend policy.},label={lst:div}]
@
\sethlcolor{humanpurple}  
\hl{
Query:
}
@  What is Microsoft's dividend policy and how does it compare to other companies in the technology sector?

@
\sethlcolor{cyan}  
\hl{
WebGLM:
}
@ Microsoft has a history of paying dividends to its shareholders. The company's most recent dividend increase was a 10$\%$ increase on November 17, 2021, which brought its per-share payout to $\$$0.68. Microsoft has also yields 0.9$\%$ on its stock. In comparison to other technology companies, Microsoft's dividend policy is considered to be a good one.

// citations to information sources 
More details: 1. [Investor's Business Daily], 2. [The Motley Fool], 3. [Microsoft], 4. [Forbes], 5. [Barron's].

@
\sethlcolor{aigold}  
\hl{
FinGPT:
}
@ Microsoft has a history of paying dividends to its shareholders. The company's most recent dividend increase was a 10$\%$ increase on November 17, 2021, which brought its per-share payout to $\$$0.68. Microsoft has also yields 0.9$\%$ on its stock. In comparison to other technology companies, Microsoft's dividend policy is considered to be a good one.

@
\sethlcolor{aired}  
\hl{
FinChat:
}
@ Microsoft Corporation has a dividend policy in place. As of the most recent data, Microsoft pays a dividend four times a year. The dividend amount is $\$$3.00 per share. The dividend yield, which is the dividend amount divided by the stock price, is 0.88$\%$ as of October 25, 2023.

Microsoft has a consistent history of paying dividends, with the first dividend payment dating back to February 19, 2003. The most recent dividend payment was made on November 15, 2023. In terms of how Microsoft's dividend policy compares to other companies in the technology sector, it is important to note that dividend policies can vary among companies...(long details omitted).

@
\sethlcolor{aigreen}  
\hl{
WeaverBird:
}
@  Microsoft's dividend policy is to pay a quarterly dividend of $\$$0.68 per share. This means that for each share of Microsoft stock, its shareholders will receive $\$$0.68 every 3 months or 4 times per year. The company's dividend payments are typically made in the form of a quarterly dividend, which is a payment made by the company to its shareholders on a regular basis. When comparing Microsoft's dividend policy to other companies in the technology sector, it is important to consider the company's payout ratio, which is the percentage of its earnings that are paid out to shareholders as dividends. Microsoft's payout ratio is 66.1$\%$, which is relatively high compared to some other technology companies. For example, Apple, Microsoft's main competitor in the technology sector, pays a dividend of only 3.4$\%$ of its earnings out to shareholders. Google's dividend policy is similar to Microsoft's, but its payout ratio is lower at 52.5$\%$.

Overall, Microsoft's dividend policy is relatively high compared to other technology companies, and its shareholders can expect to receive regular dividends as the company continues to perform well.

// citations to information sources 
More details: 1. [Dividends Diversify], 2. [The Motley Fool], 3. [Seeking Alpha], 4. [Forbes], 5. [Barron's].
\end{lstlisting}

The second example is in Chinese. 
As we can see, our WeaverBird system exhibits superior performance in answering Chinese finance-related questions, which neither WebGLM nor FinChat could answer well. \\

\begin{lstlisting}[caption={A Chinese example.},label={lst:chinese}]
@
\sethlcolor{humanpurple}  
\hl{
Query:
}
@ @\begin{CJK}{UTF8}{gbsn}半夏资本李蓓的最新投资观点是什么?\end{CJK}@

<-- English translation
What is the latest investment viewpoint of Li Bei at Banxia Capital?
-->

@
\sethlcolor{cyan}  
\hl{
WebGLM:
}
@ @\begin{CJK}{UTF8}{gbsn}半夏资本李蓓近日在半夏投资公众号发表了两篇文章，对当前全球宏观形势和投资机会进行了展望。她表示，人民币和中国股市或进入牛市。在今年的4月底，她曾经提出过“股票风险已经越来越小”的观点。半夏投资是一家极具流量的知名私募公司，李蓓经常通过官方公众号发布自己对投资、市场以及一些热点话题的看法，且观点鲜明，行文犀利。\end{CJK}@

<-- English translation
Li Bei, from Banxia Capital, recently published two articles on the Banxia Investment official WeChat account, providing insights into the current global macro situation and investment opportunities. She mentioned that the Renminbi and Chinese stock market may enter a bull market. In late April of this year, she also expressed the view that "stock market risks are becoming smaller". Banxia Investment is a well-known private equity firm with a large following. Li Bei often shares her views on investment, markets, and hot topics through the official WeChat account, presenting distinct opinions and sharp writing style.
-->

More Details: 1. [stcn.com], 2. [stcn.com], 3. [ce.cn], 4. [sina.cn], 5. [sina.com.cn].

@
\sethlcolor{aigold}  
\hl{
FinGPT:
}
@ @\begin{CJK}{UTF8}{gbsn} 半夏资本投资总监李蓓在投资界活动分享中表示，A股市场投资难度越来越大，价值投资和成长投资之间的边界越来越模糊，在这样的市场环境中，投资策略的边界也变得越来越模糊。她认为，在这样的市场环境中，投资策略的边界也变得越来越模糊，在这样的市场环境中，投资者需要有更加开放的心态，能够灵活应对市场变化，寻找一些未被充分发掘的投资机会。\end{CJK}@

<-- English translation
Banxia Capital's Investment Director, Li Bei, shared during an investment conference that investing in the A-share market has become increasingly challenging, with the boundaries between value investing and growth investing becoming more blurred. In such a market environment, the boundaries of investment strategies have also become increasingly unclear. She believes that in such a market environment, investors need to have a more open mindset and be flexible in responding to market changes, seeking out investment opportunities that have not been fully explored.
-->

@
\sethlcolor{aired}  
\hl{
FinChat:
}
@ @I'm sorry, but I couldn't find any information about\begin{CJK}{UTF8}{gbsn}半夏资本李蓓\end{CJK}@'s latest investment views in the provided data sources. The data sources primarily focus on different companies and their financial information.

<-- English translation
Based on the provided data, there is no information available about the latest investment views of Li Bei at Banxia Capital. The data source does not include Li Bei's investment perspectives. If you have any other questions about Li Bei at Banxia Capital, I will do my best to answer them.
-->

@
\sethlcolor{aigreen}  
\hl{
WeaverBird:
}
@ @\begin{CJK}{UTF8}{gbsn}根据已知信息，半夏投资创始合伙人兼基金经理李蓓在最近的采访中透露了一些最新的投资观点。她表示，过去一个月，半夏投资在持续提高股票仓位，主要集中在中下游大盘股。另外，她也提到了跨资产类别的好处，认为当你发现一个大的经济趋势的时候，并不是说投股票才是最好的选择，有些时候可能商品是更好的选择。李蓓还表示，基金规模过大并不利于宏观对冲策略的发挥。\end{CJK}@

<-- English translation
Based on the known information, Li Bei, the co-founder and fund manager of Banxia Capital, revealed some of the latest investment perspectives in a recent interview. She mentioned that over the past month, Banxia Capital has been consistently increasing its stock positions, mainly focusing on large-cap stocks in the mid-to-low sectors. Additionally, she emphasized the benefits of diversifying across asset classes, stating that when you identify a major economic trend, it doesn't always mean investing in stocks is the best choice; sometimes commodities might be a better option. Li Bei also mentioned that having an excessively large fund size is not conducive to the implementation of macro hedging strategies.
-->

More Details: 1. [21jingji.com], 2. [stcn.com], 3. [ce.cn], 4. [sina.cn], 5. [sina.com.cn].
\end{lstlisting}

\vspace{3mm}
\section{Conclusion}
\label{sec:conclusion}
We presented an open-source, intelligent dialogue system tailored for the finance sector, which outperforms the best available solutions as evidenced by its superior capabilities in solving a wide range of finance-related questions.
Our systematic approach contributes to the line of research on building domain-specific large language models. In addition, our training and inference strategies may be useful for developing retrieval-based dialogue systems in other domains, allowing us to unlock broader real applications. 



\vspace{3mm}
\bibliographystyle{ACM-Reference-Format}
\bibliography{hongyu_nlp}

\appendix

\vspace{3mm}

\section{Ongoing and Future Work}\label{app:future_work}
We are currently exploring several extensions to deal with more complex dialogue cases in the finance domain. We are particularly interested in handling:
\begin{itemize}[leftmargin=*]
\item Multiple-turn conversations. 
In such scenarios, the system has to engage in the communication and provide new responses at each turn that are consistent and coherent with previous statements. 

\item External tools.  
Users may want our system to interact with other external tools, such as pulling information from commercial data vendors (e.g., Bloomberg Terminal) and placing transaction orders in online brokerage software (e.g., Charles Schwab). 

\item More user-friendly presentation. 
Users may desire our system presenting answers in richer formats such as tables and diagrams. 
\end{itemize}
\vspace{3mm}
\section{Software Interface}\label{app:software}
\begin{figure*}[t]
    \includegraphics[width=0.99\linewidth]{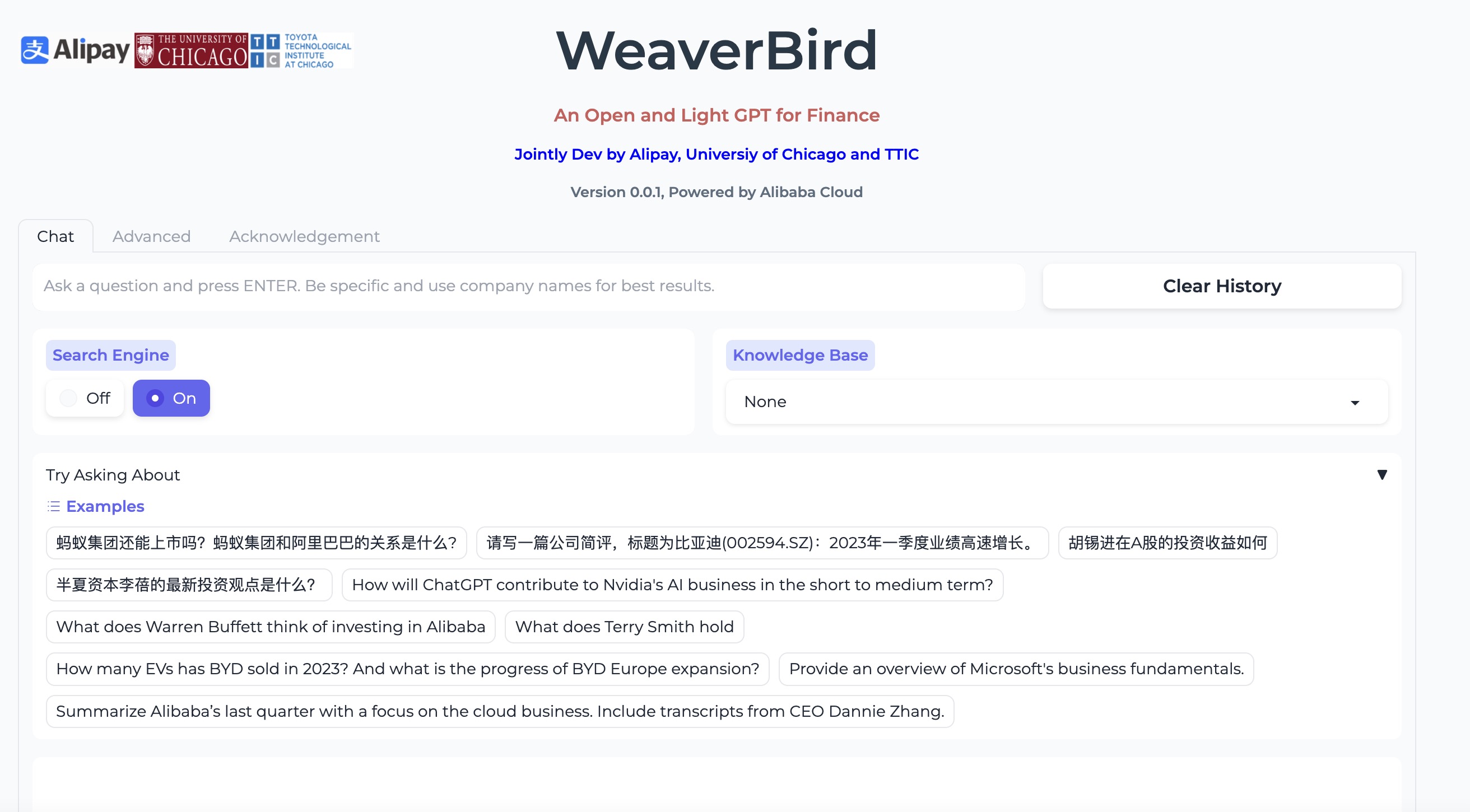}
    \caption{The main interface of WeaverBird: the configuration and chatbox.}
    \label{fig:webui}
\end{figure*}

\begin{figure*}[t]
    \includegraphics[width=0.99\linewidth]{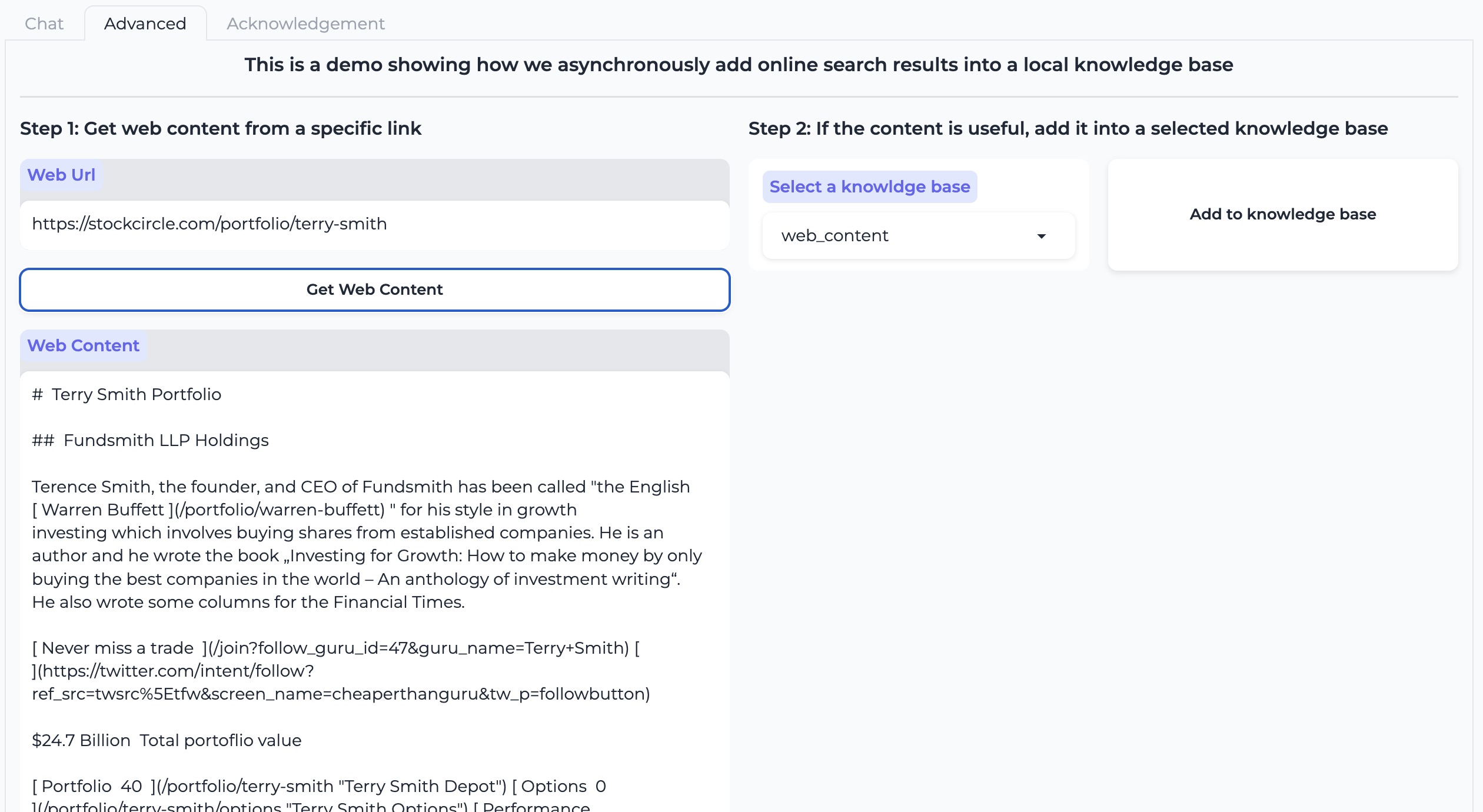}
    \caption{The 'advanced' tab of WeaverBird: the user can choose to save useful search results into a local knowledge base.}
    \label{fig:webui2}
\end{figure*}

The main interface of our WeaverBird system can be seen in \cref{fig:webui,fig:webui2}. Experience it at {\small\url{https://weaverbird.ttic.edu}}. 

\end{document}